\documentclass[10pt,twocolumn,letterpaper]{article}

\usepackage{cvpr}
\usepackage{times}
\usepackage{epsfig}
\usepackage{graphicx}
\usepackage{amsmath}
\usepackage{amssymb}

\usepackage[utf8]{inputenc} 
\usepackage{booktabs}       
\usepackage{amsfonts}       
\usepackage{nicefrac}       
\usepackage{microtype}      
\usepackage{algorithm}
\usepackage{algpseudocode}
\usepackage{multirow}
\usepackage{mathtools}
\DeclarePairedDelimiter{\norm}{\lVert}{\rVert}

\usepackage[breaklinks=true,bookmarks=false]{hyperref}

\cvprfinalcopy 

\def\cvprPaperID{10141} 

\ifcvprfinal\fi
\begin{document}

\title{Detecting Adversarial Samples Using Influence Functions and Nearest Neighbors}

\author{
  Gilad Cohen \\
  Tel Aviv University \\
  Tel Aviv, 69978 \\
  {\tt\small giladco1@mail.tau.ac.il} \\
  \and
  Guillermo Sapiro \\
  Duke University \\
  North Carolina, 27708 \\
  {\tt\small guillermo.sapiro@duke.edu} \\
  \and
  Raja Giryes \\
  Tel Aviv University \\
  Tel Aviv, 69978 \\
  {\tt\small raja@tauex.tau.ac.il}
}

\maketitle
\thispagestyle{empty}

\begin{abstract}
Deep neural networks (DNNs) are notorious for their vulnerability to adversarial attacks, which are small perturbations added to their input images to mislead their prediction. Detection of adversarial examples is, therefore, a fundamental requirement for robust classification frameworks. In this work, we present a method for detecting such adversarial attacks, which is suitable for any pre-trained neural network classifier. We use influence functions to measure the impact of every training sample on the validation set data. From the influence scores, we find the most supportive training samples for any given validation example. A $k$-nearest neighbor ($k$-NN) model fitted on the DNN's activation layers is employed to search for the ranking of these supporting training samples.
We observe that these samples are highly correlated with the nearest neighbors of the normal inputs, while this correlation is much weaker for adversarial inputs.
We train an adversarial detector using the $k$-NN ranks and distances and show that it successfully distinguishes adversarial examples, getting state-of-the-art results on six attack methods with three datasets. Code is available at \url{https://github.com/giladcohen/NNIF_adv_defense}.
\end{abstract}

\section{Introduction}
\label{Introduction}

\makeatletter
\setlength{\@fptop}{0pt}
\makeatother
\begin{figure}[t!]
\centering
\includegraphics[width=\linewidth]{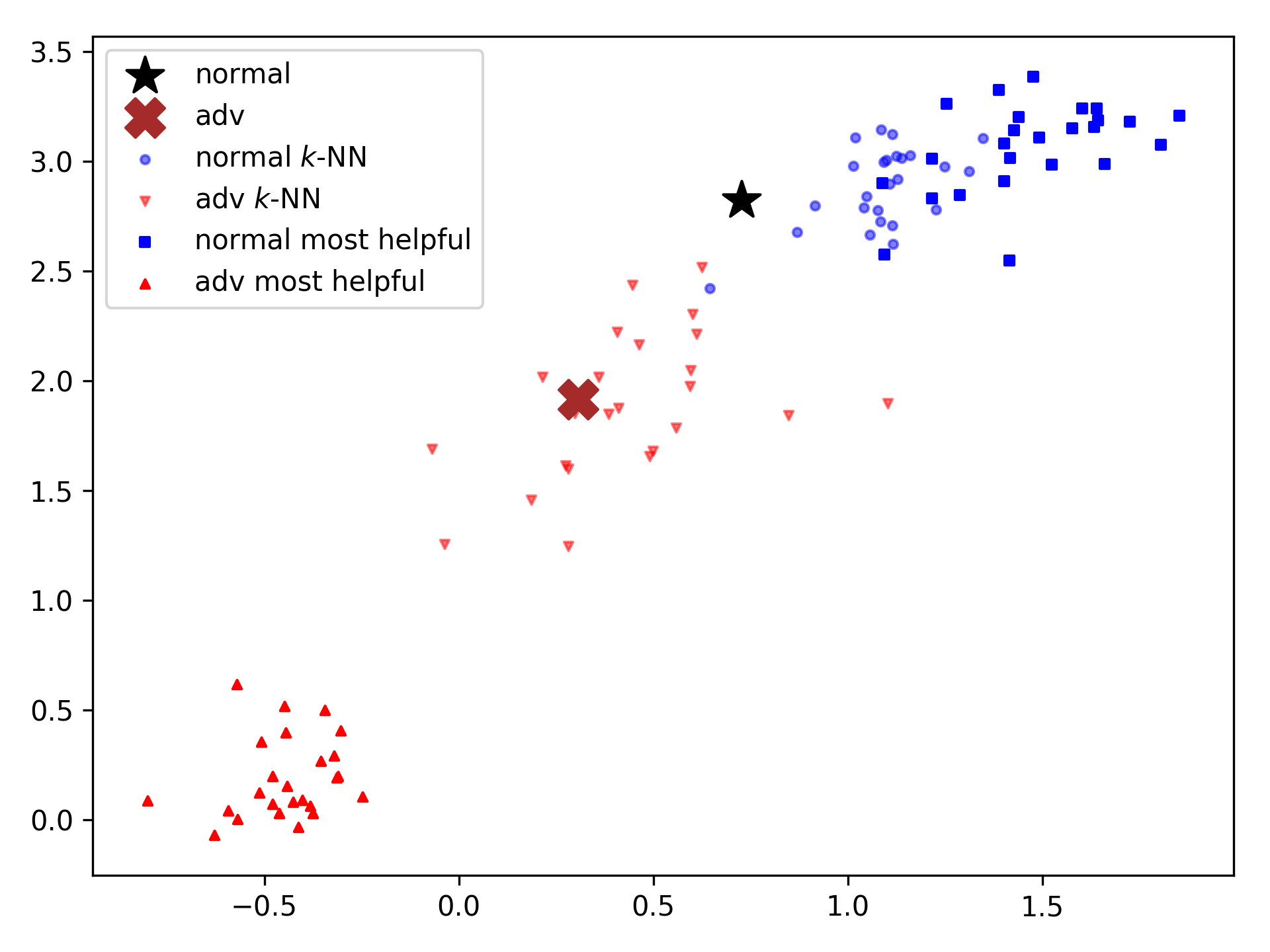}
\caption{The correspondence between the  helpful examples based on influence functions and the $k$-nearest neighbours ($k$-NN) in the embedding space of a DNN can help to distinguish adversarial examples from normal ones. We present (using PCA) the embedding space of a DNN for a normal example (black star) with its adversarial version (brown \textbf{X}) along with their $k$-NN ($k$=25) and 25 most helpful samples. Note that for the normal example, the helpful samples highly correlate with the $k$-NN in the embedding space. Yet, in the adversarial case, these samples are far from each other. This observation leads us to a technique for detecting adversarial attacks.}
\label{images/teaser}
\end{figure}

Deep Neural Networks (DNNs) are vastly employed in both the academy and industry, achieving state-of-the-art (SOTA) results in many domains such as computer vision \cite{Krizhevsky2012ImageNetNetworks,Schroff2015FaceNet:Clustering,Voulodimos2018DeepReview}, natural language processing \cite{DBLP:journals/corr/BahdanauCB14,Kim2014ConvolutionalClassification}, and speech recognition \cite{Hinton2012DeepRecognition,DBLP:journals/corr/ZhangPBZLBC17}. However, studies have shown that DNNs are vulnerable to adversarial examples \cite{FGSM,Intriguing}, which are specially crafted perturbations on their input. Adversarial attacks generate such examples that fool machine learning models, inducing them to predict erroneously with high confidence, while being imperceptible to humans. Adversarial subspaces of different DNN classifiers tend to overlap, which makes some adversarial examples generated for a surrogate model fool also other different unseen DNNs \cite{Transferable}. This makes adversarial attacks a real threat to any machine learning model and thus should be kept in mind while deploying a DNN.

The vulnerability of neural networks puts into question their usage in sensitive applications, where an opponent may provide modified inputs to cause misidentifications. For this reason, many methods have been developed to face this challenge. They can be mainly divided into two groups: 1) \textit{proactive} defense methods, which aim at improving the robustness of DNNs to adversarial examples, and 2) \textit{reactive} detection techniques that do not change the DNN but rather try to find whether an attack is associated with a certain input or not. 

{\bf Contribution.} In this work, we focus on the reactive detection problem. We propose a novel strategy for detecting adversarial attacks that can be applied to any pre-trained neural network.
The core idea of the algorithm is that there should be a correspondence between the training data and the classification of the network. If this relationship breaks then it is very likely that we are in the case of an adversarial input. 

To this end, we use two ``metrics'' to check the impact of the training data on the network decision. The first is influence functions \cite{Koh2017UnderstandingBP}, which determines how data points in the training set influence the decision of the network for a given test sample. This metric measures how much a small upweighting of a specific training point in the model's loss function affects the loss of a testing point. Thus, it provides us with a measure of how much a test sample classification is affected by each training sample.

Second, we apply a $k$-nearest neighbor ($k$-NN) classifier at the embedding space of the network. Various recent works \cite{Deep_kNN_Papernot,Doring2017_knn_convergence,to_trust_or_not_a_classifier,DNN_or_kNN} demonstrate a high correlation between the network softmax output and the decision of a $k$-NN applied at the embedding space of this network (where the neighbors are chosen from the training set). They basically show that the network's decision relies on the nearest neighbors resemblance in the embedding space. Thus, the distance in that space may serve as a measure for the effect of an example on the network output. 

Given the influence function and $k$-NN based measures, we turn to combine them together to generate a novel strategy to detect adversarial examples. The rationale behind our approach is that for a normal input, its $k$-NN training samples (nearest neighbors in the embedding space) and the most helpful training samples (found using the influence function) should correlate. Yet, for adversarial examples this correlation should break and thus, it will serve as an indication that an attack is happening.

Figure~\ref{images/teaser} illustrates this relationship between the $k$-NN and the most helpful training samples. The black star and brown \textbf{X} denote a normal and its corresponding adversarial image from CIFAR-10 validation set; the plot is of the embedding space projected using PCA fitted on the training set. For each sample (normal/adv), we find its 25 nearest neighbors (blue circles/red downward triangles) in the DNN embedding space; in addition, we find its 25 most helpful training examples from the training set (marked as blue squares and red upward triangles, respectively). Note that the nearest neighbors and the top most helpful training samples of the normal image are very close in the PCA embedding space, whereas the adversarial image does not exhibit the same correspondence between the training samples.

To check the correlation between the two, we pursue the following strategy: For an unseen input sample, we take the most influential examples from the training set chosen by the influence functions. Then, we check their distance ranking in the embedding space of the network (i.e., what value of $k$ will cause $k$-NN to take them into account) and their $L_2$ distance from the input sample's embedding vector.
Finally, we use these $k$-NN features to train a simple Logistic Regression (LR) for detecting whether the input is adversarial or not.

We evaluate our detection strategy on various attack methods and datasets showing its advantage over other leading detection techniques. The results confirm the hypothesis claimed in previous works on the resemblance between $k$-NN applied on the embedding space and the DNN decision, and show how it can be used for detecting adversarial examples.

\section{Related work}
\label{Related work}

In this section, we briefly review existing  papers on adversarial attacks and defenses, and related theory.

\textbf{Theory:} Madry et al. used the framework of robust optimization and showed results of adversarial training \cite{Towards_DeepLearning_Resistance}. They found that projected gradient descent (PGD) is an optimal first order adversary, and employing it in the DNN training leads to optimal robustness against any first order attack. Simon-Gabriel et al. demonstrated that DNNs' vulnerability to adversarial attacks is increased with the gradient of the training loss as a function of the inputs \cite{Vulnerability_Input_Dimension}. They also found that this vulnerability does not depend on the DNN model.

Fawzi et al. studied the geometry and complexity of the functions learned by DNNs and provided empirical analysis of the curvatures of their decision boundaries \cite{ClassificationRegions}. They showed that a DNN classifier is most vulnerable where its decision boundary is positively curved and that natural images are usually located in the vicinity of flat decision boundaries. These findings are also supported by Moosavi-Dezfooli et al. \cite{Curvature}, who found that positively curved decision boundaries increase the likelihood that a small universal perturbation would fool a DNN classifier.

Some works provided guarantees to certify robustness of the network. Hein and Andriushchenko formalized a formal upper bound for the noise required to flip a network prediction \cite{Cross_Lipschitz_regularization}, while Sinha et al. provided an efficient and fast guarantee of robustness for the worst-case population performance, with high probability \cite{CertifyingRobustness}.

\textbf{Adversarial attacks:} One of the simplest and fastest attack methods is the fast gradient sign method (FGSM) \cite{FGSM}; in this method the attacker linearly fits the cross entropy loss around the attacked sample and lightly perturbs the image pixels in the direction of the gradient loss. This is a fast one-step attack, which is very easy to deploy on raw input images.

The Jacobian-based saliency map attack (JSMA) \cite{JSMA} takes a different approach. Instead of mildly changing all image pixels, this attack is crafted on the $L_0$ norm, finding one or two pixels which induce the largest change in the loss and modify only them. This is a strong attack, achieving 97\% success rate by modifying only 4.02\% of the input features on average. Yet, it is iterative and costly.

Deepfool \cite{DeepFool} proposed by Moosavi-Dezfooli et al. is a non-targeted attack\footnote{Non-targeted attacks are adversarial attacks which aim to make the prediction incorrect regardless of the spricifc erroneous class.} that creates an adversarial example by moving the attacked input sample to its closest decision boundary, assuming an affine classifier.
In reality most DNNs are very non linear, however, the authors used an iterative method, linearizing the classifier around the test sample at every iteration. Compared to FGSM and JSMA, Deepfool performs less perturbations to the input. It was also employed in the Universal Perturbations attack \cite{Dezfooli17Universal}, which is an iterative attack that aims at fooling a group of images using the same minimal, universal perturbation applied on all of them.

Carlini and Wagner \cite{CarliniWagner2017Towards} proposed a targeted attack\footnote{Targeted attacks are adversarial attacks which aim to make the prediction classified to a particular erroneous class.} (denoted as CW) to impact the defensive distillation method \cite{Distillation}. The CW attack is resilient against most adversarial detection methods.
In another work Carlini and Wagner provided an optimization framework \cite{Carlini2017BypassingTen}, which includes a defense-specific loss as a regularization term. This optimization-based attack is argued to be the most effective to date for a white-box threat model where the adversary knows everything related to the trained DNN: training data, architecture, hyper-parameters, weights, etc. Chen et al. \cite{EAD} included a $L_1$ regularization to the CW attack, forming the Elastic-net Attack to DNNs (EAD).

\textbf{Adversarial defenses:} A wide range of proactive defense approaches have been proposed, including adversarial (re)training \cite{FGSM,Kurakin2017AdversarialML,tramer2018ensemble,Shaham2018RobustOptimization,Virtual_Adversarial_Training}, distillation networks \cite{Distillation}, gradient masking \cite{tramer2018ensemble}, feature squeezing \cite{Xu2018FeatureSD}, network input regularization \cite{InputGradients,Jakubovitz2018ImprovingDR}, output regularization \cite{Cross_Lipschitz_regularization}, adjusting weights of correctly predicted labels \cite{BANG}, Parseval networks \cite{Parseval}, and $k$-NN search \cite{Dubey2019DefenseAA,Sitawarin2019DefendingAA}.

However, those defenses can be evaded by the optimization-based attack \cite{Carlini2017BypassingTen}, either wholly or partially. Since there are no known intrinsic properties that differentiate adversarial samples from regular images, proactive adversarial defense is extremely challenging. Instead, recent works have focused on reactive adversarial detection methods, which aim at distinguishing adversarial images from natural images, based on features extracted from DNN layers \cite{Metzen17detecting,Li2017AdversarialED,Rouhani2018TowardsSD} or from a learned encoder \cite{Meng2017MagNetAT}.
Feinman et al. \cite{Feinman2017DetectingAS} proposed a LR detector based on Kernel density (KD) and Bayesian uncertainty features.

Ma et al. \cite{LID} characterized the dimensional properties of the adversarial subspaces regions and proposed to use a property called Local Intrinsic Dimentionaloty (LID) . LID describes the rate of expansion in the number of data objects as the distance from the reference sample increases. The authors estimated the LID score at every DNN layer using extreme value theory, where the smallest NN distances are considered as extreme events associated with the lower tail of the data samples' underlying distance distribution. Given a pretrained network and a dataset of normal examples, the authors applied on every sample: 1) Adversarial attack. 2) Addition of Gaussian Noise.
The natural and noisy images were considered as negative (non-adversarial) class and the adversarial images were considered as positive class. For each image (natural/noisy/adversarial) they calculated a LID score at every DNN layer. Lastly, a LR model was fitted on the LID features for the adversarial detection task.

Papernot and McDaniel \cite{Deep_kNN_Papernot} proposed the Deep $k$-Nearest Neighbors (D$k$NN) algorithm to estimate better the prediction, confidence, and credibility for a given test sample. Using a pretrained network, they fitted a $k$-NN model at every layer. Next, they used a left-out calibration set to estimate the nonconformity of every test sample for label j, counting the number of nearest neighbors along the DNN layer which differs from j. They showed that when an adversarial attack is made on a test sample, the real label displays less correspondence with the $k$-NN labels from the DNN activations along the layers.

Lee et al. \cite{Mahalanobis_adv_detection} trained generative classifiers using the DNN activations of the training set on every layer to detect adversarial examples by applying a Mahalanobis distance-based confidence score. First, for every class and every layer, they computed the empirical mean and covariance of the activations induced by the training samples. Next, using the above class-conditional Gaussian distributions, they calculated the Mahalanobis distance between a test sample and its nearest class-conditional Gaussian. These distances are used as features to train a LR classifier for the adversarial detection task. The authors claimed that using the Mahalanobis distance is significantly more effective than the Euclidean distance employed in \cite{LID} and showed improved detection results.

\section{Method}
\label{Method}
We hypothesize that the DNN predictions are influenced by the $k$-NN of the training data in their hidden layers, especially in the embedding layer. 
If so, in order to fool the network, an adversarial attack must move the test sample towards a "bad" subspace in the embedding space, where harmful training data can cause the network to misclassify the correct label. To inspect our hypothesis, we fitted a $k$-NN model on the DNN's activation layers, and also employed the influence functions as used in \cite{Koh2017UnderstandingBP}.


Influence functions can interpret a DNN by pointing out which of the training samples helped the DNN to make its prediction, and which training samples were harmful, i.e., inhibited the network from its  prediction.
Koh and Liang \cite{Koh2017UnderstandingBP} suggested to measure the influence a train image $z$ has on the loss of a specific test image $z_{test}$, by the term:
\begin{equation}
\label{I_up_loss}
I_{up,loss}(z, z_{test}) = -\nabla_{\theta}L(z_{test}, \theta)^TH_{\theta}^{-1}\nabla_{\theta}L(z, \theta),
\end{equation}
where $H$ is the Hessian of the machine learning model, $L$ is its loss, and $\theta$ are the model parameters. In the definition of Eq.~\eqref{I_up_loss} $z$ and $z_{test}$ are images.

For each test example $z_{test}$, we calculate Eq.~\eqref{I_up_loss} per each training example $z$ in the training set. Then, we sort all $I_{up,loss}(z, z_{test})$ scores, determining the top $M$ helpful and harmful training examples for a specific $z_{test}$. Next, for each of the 2x$M$ selected training points we find its rank and distance from the testing example by fitting a $k$-NN model on the embedding space using all the training examples' embedding vectors. We feed the embedding vector of each test sample $z_{test}$ to the $k$-NN model to extract all the nearest neighbors' ranks (denoted as $\mathcal{R}$) and distances (denoted $\mathcal{D}$) of the examples in the training set. The $\mathcal{R}$ and $\mathcal{D}$ features can also be extracted from any other hidden activation layer within the DNN, and not solely from the embedding vector.
$\mathcal{R}^{M\uparrow}$, $\mathcal{D}^{M\uparrow}$ and $\mathcal{R}^{M\downarrow}$, $\mathcal{D}^{M\downarrow}$ are all the ranks and distances of the helpful and harmful training examples, respectively.

We apply an adversarial attack on $z_{test}$ and repeat the aforementioned process on the new, crafted image. Both the normal and adversarial features ($\mathcal{R}^{M\uparrow}$, $\mathcal{D}^{M\uparrow}$,  $\mathcal{R}^{M\downarrow}$, $\mathcal{D}^{M\downarrow}$) are used to train a LR classifier for the adversarial detection task. The detector training scheme is described in Alg.~\ref{alg:NNIF}. 

\begin{algorithm*}[h!]
\caption{Adversarial detection using Nearest Neighbors Influence Functions (NNIF)}\label{alg:NNIF}
\begin{algorithmic}[1]
\Require Training set ($X_{train}$, $Y_{train}$) and validation set ($X_{val}$, $Y_{val}$)
\Require Pre-trained DNN with L activation layers and parameters $\theta$ 
\Require $M$: Number of top influence samples to collect
\Ensure Detector($\mathcal{R}^{M\uparrow}$,  $\mathcal{D}^{M\uparrow}$, $\mathcal{R}^{M\downarrow}$,  $\mathcal{D}^{M\downarrow}$)  \Comment{An adversarial example detector}

\item[]
\State $N_{train} = |X_{train}|$, $N_{val} = |X_{val}|$ \Comment{Number of examples in train- and validation-set}

\State Initialize: $R_{norm}^+$=[], $D_{norm}^+$=[], $R_{norm}^-$=[], $D_{norm}^-$=[] \Comment{Normal image features}
\State Initialize: $R_{adv}^+$=[], $D_{adv}^+$=[], $R_{adv}^-$=[], $D_{adv}^-$=[] \Comment{Adversarial image features}
\State ($X_{val}^{adv}, Y_{val}^{adv}$) := adversarial attack on ($X_{val}$, $Y_{val}$) \Comment{Generate a new adversarial dataset by attacking the validation set}

\For{{\textit{l} in [1,L]}}
  \State Fit $k$-NN[\textit{l}] model on layer \textit{l}. $k=N_{train}$ 
  \For{($x_i, y_i$) in ($X_{val}$, $Y_{val}$)}
    \State $\mathcal{R}^{M\uparrow}$,  $\mathcal{D}^{M\uparrow}$, $\mathcal{R}^{M\downarrow}$,  $\mathcal{D}^{M\downarrow}$ := \textsc{NNFeatures}($x_i$, $k\text{-NN}$[\textit{l}]) \Comment{Get NNIF helpful/harmful features for normal images}
    \State $R_{norm}^+$.append($\mathcal{R}^{M\uparrow}$), $D_{norm}^+$.append($\mathcal{D}^{M\uparrow}$),  $R_{norm}^-$.append($\mathcal{R}^{M\downarrow}$), $D_{norm}^-$.append($\mathcal{D}^{M\downarrow}$)
  \EndFor
  \For{($x_i, y_i$) in ($X_{val}^{adv}$, $Y_{val}^{adv}$)}
    \State $\mathcal{R}^{M\uparrow}$,  $\mathcal{D}^{M\uparrow}$, $\mathcal{R}^{M\downarrow}$, $\mathcal{D}^{M\downarrow}$ := \textsc{NNFeatures}($x_i$, $k\text{-NN}$[\textit{l}]) \Comment{Get NNIF helpful/harmful features for adv images}
    \State $R_{adv}^+$.append($\mathcal{R}^{M\uparrow}$), $D_{adv}^+$.append($\mathcal{D}^{M\uparrow}$),  $R_{adv}^-$.append($\mathcal{R}^{M\downarrow}$), $D_{adv}^-$.append($\mathcal{D}^{M\downarrow}$)
  \EndFor
\EndFor

\State $NNIF_{pos}$ = ($R_{adv}^+$, $D_{adv}^+$, $R_{adv}^-$, $D_{adv}^-$)
\State $NNIF_{neg}$ = ($R_{norm}^+$, $D_{norm}^+$, $R_{norm}^-$, $D_{norm}^-$)

\State Detector($\mathcal{R}^{M\uparrow}$,  $\mathcal{D}^{M\uparrow}$, $\mathcal{R}^{M\downarrow}$,  $\mathcal{D}^{M\downarrow}$) = train a classifier on ($NNIF_{pos}$, $NNIF_{neg}$)
 
\item[]
\Procedure{NNFeatures}{$x_i$, $k$-NN[\textit{l}]} \Comment{Collecting nearest neighbors features}
  \State Initialize: $R^+$=[], $D^+$=[], $R^-$=[], $D^-$=[] \Comment{image's nearest neighbors features}

  \State $\mathcal{R}$, $\mathcal{D}$ := Apply $k\text{-NN}$ on activation layer \textit{l}, get training examples' ranks \& $L_2$ distances out of activations of sample $x_i$
  \State $H_{inds}^+$, $H_{inds}^-$ := $\textsc{InfluenceFunction}((x_i, y_i), (X_{train}, Y_{train}))$ \Comment{get indices of the 2x$M$ most influencing training samples. This procedure is presented in the supp. material.}
  \For{j in $H_{inds}^+$} \Comment{Collect M helpful ranks and distances}
    \State $R^+$.append($\mathcal{R}[j]$)
      \State $D^+$.append($\mathcal{D}[j]$)
  \EndFor
  \For{j in $H_{inds}^-$} \Comment{Collect M harmful ranks and distances}
    \State $R^-$.append($\mathcal{R}[j]$)
    \State $D^-$.append($\mathcal{D}[j]$)
  \EndFor
  \State \textbf{return} $R^+$, $D^+$, $R^-$, $D^-$
\EndProcedure
\end{algorithmic}
\end{algorithm*}

We name our adversarial detection method Nearest Neighbor Influence Functions (NNIF). We assume that the training, validation, and testing sets are not contaminated with adversarial examples, as in \cite{Carlini2017BypassingTen}. We start by generating an adversarial validation set from the normal validation set (step 4). The $M$ most helpful and harmful training examples associated with the validation image prediction (either normal or adversarial) are found using the influence function in step 22 (see supp. material for the \textsc{InfluenceFunction} procedure). The NNIF features are then evaluated by the $k$-NN model, extracting the ranks and distances (from $\mathcal{R}$ and $\mathcal{D}$) of the most influential training points found above. This is done for both the normal validation images (step 8) and for the adversarial images (step 12). This scheme can be carried out on the embedding layer alone, or employed for all $L$ activation layers within the DNN.

Finally, a LR classifier is trained using the NNIF features.
Images from the test set are classified to either adversarial (positive) or normal (negative) based on the NNIF features extracted from the $M$ most helpful/harmful training examples, ($\mathcal{R}^{M\uparrow}$, $\mathcal{D}^{M\uparrow}$,  $\mathcal{R}^{M\downarrow}$, $\mathcal{D}^{M\downarrow}$).

Training our NNIF detector is very time consuming, requiring us to calculate Eq.~\eqref{I_up_loss} on the entire training set for every validation image, having a time complexity of $\mathcal{O}(N_{train} \cdot N_{val})$, where $N_{train}$ and $N_{val}$ are the size of the training and validation sets, respectively. For an adversarial detection the complexity time is $\mathcal{O}(N_{train})$, since we need to find the top $M$ helpful/harmful training examples for every new incoming test image.

Papernot and McDaniel \cite{Deep_kNN_Papernot} focused on improving credibility and robustness in DNN. They used the nearest neighbors in the activation layers for interpretability. As an additional competing strategy, we convert their original D$k$NN algorithm \cite{Deep_kNN_Papernot} to an adversarial detection method. This is done by collecting the empirical p-values calculated in the D$k$NN strategy and formulating a reactive adversarial detector by training a LR model on these features.
While NNIF also uses nearest neighbors, instead of inspecting the labels of the nearest neighbors, we examine the correlation between them and the image's most helpful/harmful training examples using the influence functions.

\section{Results}

\begin{figure*}[h!]
\centering
\includegraphics[width=\linewidth]{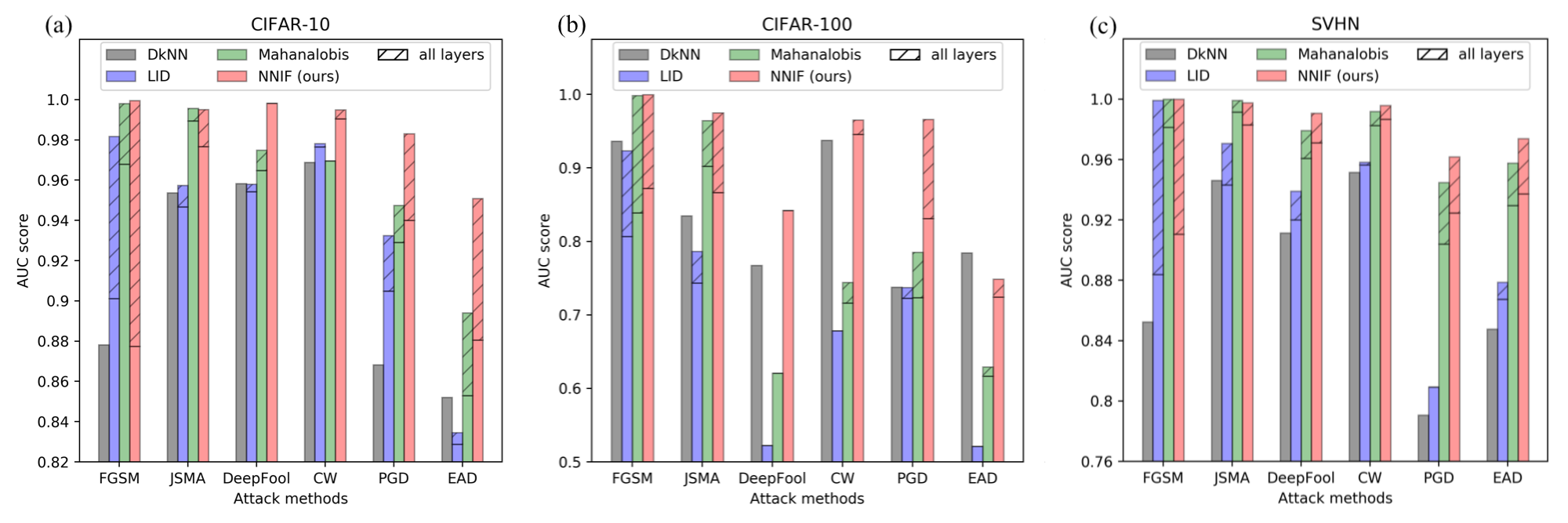}
\caption{Comparison of AUC scores for detection of FGSM, JSMA, Deepfool, CW, PGD, and EAD attacks on three datasets: (a) CIFAR-10, (b) CIFAR-100, and (c) SVHN. The black, blue, green, and red bars correspond to the D$k$NN, LID, Mahalanobis, and NNIF defense methods, respectively. The hatched pattern bars correspond to AUC scores increase where taking into consideration all the DNN activation layers instead of just the penultimate activation layer. Each attack cluster of bars is divided to four columns which correspond to the methods (from left to right): D$k$NN, LID, Mahalanobis, and NNIF. Our NNIF detector surpasses previous SOTA methods by a large margin for most of the attacks.}
\label{images/all_defenses_small}
\end{figure*}

This section shows the power of our NNIF adversarial detector against six adversarial attack strategies (norms in parentheses): FGSM ($L_\infty$), JSMA ($L_0$), DeepFool ($L_2$), CW ($L_2$), PGD ($L_\infty$), and EAD ($L_1$), as introduced in Section~\ref{Related work}. The PGD attack was used with input perturbation as implemented by \cite{Towards_DeepLearning_Resistance}. We selected these attacks for our experiments due to their effectiveness, diversity, and popularity.  For versatility, we used Deepfool and EAD as non-targeted attacks. PGD was  We applied these attacks on three datasets: CIFAR-10, CIFAR-100 \cite{CIFAR}, and SVHN \cite{SVHN}. NNIF performance is compared to the SOTA LID and Mahalanobis detectors (Section~\ref{Related work}) and also to the D$k$NN adversarial detector (Section~\ref{Method}). Lastly, we analyzed the robustness of NNIF against a white-box setting. Before presenting our results, we first describe the experimental setup used in our analysis.

\subsection{Experimental setup}
\textbf{Training and Testing:} Each of the three image datasets was divided into three subsets: \textit{training} set, \textit{validation} set, and \textit{testing} set, containing 49k, 1k, and 10k images respectively. Since our NNIF method is time consuming (especially the procedure \textsc{InfluenceFunction} in Alg.~\ref{alg:NNIF}), we randomly selected 49k and 1k \textit{training} and \textit{validation} samples, respectively, from the official SVHN training set and 10k testing samples from the official SVHN testing set. Any \textit{validation} or \textit{testing} image not correctly classified by the DNN was discarded.
For every image in the \textit{validation} and \textit{testing} sets, we generated adversarial examples using all six attack methods, as describe in Step 4 in Alg.~\ref{alg:NNIF}. Then, an equal number of normal and adversarial \textit{validation} images were used to train a LR classifier, which was later applied on the remaining \textit{testing} images for calculating the detectors metrics. We used the \textit{cleverhans} library \cite{cleverhans} to carry out all the adversarial attacks.

Since the D$k$NN method requires a calibration set, we randomly selected 33\% of the \textit{validation} set examples (after discarding the misclassifications) for calibrating it. Note that although Papernot and McDaniel \cite{Deep_kNN_Papernot} showed that the nearest neighbors can qualitatively detect adversarial attacks (see Fig.~7 in \cite{Deep_kNN_Papernot}), they did not formalize an adversarial detector. We employ their empirical $p$-values as features for the adversarial detection task.

\textbf{Training DNNs:} We trained all DNNs on the \textit{training} set while decaying the learning rate using the \textit{validation} set's accuracy score. All the DNNs used in our experiments are Resnet-34 \cite{RESNET} with global average pooling layer prior to the embedding space. The embedding vector was multiplied by a fully-connected layer for the logits calculation. We trained all three datasets for 200 epochs, with $L_2$ weight regularization of 0.0004, using a Stochastic Gradient Decent optimizer with momentum 0.9 and Nesterov updates. For evaluation we used the model checkpoint with the best (highest) validation accuracy on the image classification task. We follow the checklist in \cite{OnEvaluatingAdvRobustness2019} and report the full DNN validation/test accuracies for the clean models when not under attack and the attacks success rates (see supp. material).
These DNNs perform close to the SOTA and thus are sufficient for being used in an adversarial study without fine tuning \cite{Feinman2017DetectingAS}.

\textbf{Parameter tuning:} The number of neighbors ($k$) for LID and DkNN, the noise magnitude ($\epsilon$) for the Mahalanobis method, and the number of top influence samples to collect ($M$) for NNIF were chosen using nested cross validation within the \textit{validation} set, based on the AUC values of the detection ROC curve.
We tuned $k$ for DkNN using an exhaustive grid search between [10, ${N}/{\#classes}$], where N is the dataset size and $\#classes$ is the number of classes. For LID the number of nearest neighbors was tuned using a grid search over the range [10, 40) while using a minibatch size of 100 (as in \cite{LID}). For the Mahalanobis method we tuned $\epsilon$ using an exhaustive grid search in log-space between [$1E^{-5}$, $1E^{-2}$], and $M$ was tuned using a grid search over [10, 500]. The selected parameters are presented in the supp. material.

Running \textsc{InfluenceFunction} in Alg.~\ref{alg:NNIF} for an entire \textit{training} set is very slow. 
Thus, for every \textit{testing} set we randomly selected only 10k out of the 49k samples in the \textit{training} set and calculated $I_{up,loss}$ (Eq.~\eqref{I_up_loss}) just for them. Although this is a coarse approximation of the real nearest neighbors distribution in the \textit{training} set on the DNN embedding space, this approximation is sufficient for achieving new SOTA adversarial detection. We emphasize that this approximation was done only for the \textit{testing} set, and not for the \textit{validation} set.

\textbf{Activation layers:} The LID, Mahalanobis, and NNIF detectors can be trained using either features from the embedding space alone or using all the activation layers in the network. The D$k$NN detector portrays very poor results when it is applied on all the DNN's features (data not shown) and therefore, we present all the D$k$NN results  by training features from the embedding space alone.

\textbf{Threat model:} We consider two treat models, black-box and white-box settings. Unless stated otherwise, the default threat model is black-box, where the attacker is unaware that an adversarial detection is employed. In this setting, only the model's parameters are given to the adversary. In Section~\ref{Adaptive attack against NNIF} we also consider a white-box setting, where the attacker knows the model parameters, and also the adversarial detection scheme.

\subsection{Detection of adversarial attacks}
\label{Detection of adversarial attacks}
Figure~\ref{images/all_defenses_small} shows the discrimination power (AUC score) of the four inspected adversarial detectors: D$k$NN (black), LID (blue), Mahalabolis (green) and NNIF (red), on three popular datasets: CIFAR-10, CIFAR-100, and SVHN. We compare between the detection scores calculated for six adversarial attacks: FGSM, JSMA, Deepfool, CW, PGD, and EAD. The solid bars correspond to detections where only the penultimate activation layer was utilized. In some cases, considering all the layers in the DNN activations boosts the LID/Mahalanobis/NNIF scores; this is portrayed as a complementary hatched patterned bar above the solid bar.

Our method surpasses all other detectors for distinguishing Deepfool, CW, and PGD attacks, for all the datasets. On FGSM and JSMA our NNIF detector also demonstrates SOTA results, matching the Mahalanobis detector's performance. Against EAD we show new SOTA for CIFAR-10 and SVHN, but not for CIFAR-100. Table~\ref{detection_scores} summarizes the AUC scores of the detectors using features from all the DNN's activation layers. The only exception is the D$k$NN method, which is employed only on the embedding space. In the supp. material, we include results for more attacks and a similar table for the obtained AUC scores using only the DNN's penultimate layer.


\begin{table}
\centering
\caption{Comparison of AUC scores (\%) for various adversarial detection methods. Results obtained using all the DNN's activation layers for LID/Mahalanobis/NNIF and only the embedding space for D$k$NN.}
\label{detection_scores}
\resizebox{\columnwidth}{!}{
    \begin{tabular}{cc|cccccc}
    \toprule
    Dataset & Detector & FGSM & JSMA & Deepfool & CW & PGD & EAD \\
    \hline
    \multirow{4}{*}{CIFAR-10}  & D$k$NN      & 87.81          & 95.37          & 95.82          & 96.88          & 86.83 & 85.20 \\
                               & LID         & 98.18          & 95.74          & 95.80          & 97.82          & 93.24 & 83.46 \\
                               & Mahalanobis & 99.80          & \textbf{99.56} & 97.49          & 96.48          & 94.74 & 89.41 \\
                               & NNIF (ours) & \textbf{99.96} & 99.50          & \textbf{99.32} & \textbf{99.5}  & \textbf{98.31} & \textbf{95.09} \\
    \hline
    \multirow{4}{*}{CIFAR-100} & D$k$NN      & 93.65          & 83.46          & 76.71          & 93.77          & 73.78 & \textbf{78.42} \\
                               & LID         & 92.33          & 78.63          & 51.61          & 67.83          & 73.71 & 51.11 \\
                               & Mahalanobis & 99.87          & 96.44          & 62.05          & 74.43          & 78.53 & 62.93 \\
                               & NNIF (ours) & \textbf{99.96} & \textbf{97.50} & \textbf{77.17} & \textbf{96.51} & \textbf{96.60} & 74.86 \\
    \hline
    \multirow{4}{*}{SVHN}      & D$k$NN      & 85.24          & 94.61          & 91.13          & 95.15          & 79.07 & 84.77 \\
                               & LID         & 99.92          & 97.06          & 93.90          & 95.82          & 80.12 & 87.86 \\
                               & Mahalanobis & \textbf{100.00}& \textbf{99.91} & 97.92          & 99.18          & 94.47 & 95.77 \\
                               & NNIF (ours) & \textbf{100.00}& 99.76          & \textbf{99.06} & \textbf{99.59} & \textbf{96.18} & \textbf{97.40} \\
    \bottomrule
    \end{tabular}
    }
\end{table}

\subsection{Ablation study}

\begin{table}
\centering
\caption{Ablation test for adversarial attack detection: Calculating AUC score and accuracy for selected features. Attacking CIFAR-10 dataset using Deepfool.}
\label{ablation_table}
\begin{tabular}{c c c c c c}
\toprule
$\mathcal{R}^{M\uparrow}$ & $\mathcal{D}^{M\uparrow}$ & $\mathcal{R}^{M\downarrow}$ & $\mathcal{D}^{M\downarrow}$ & AUC(\%) & acc(\%) \\

\hline
& & & \checkmark & 82.11 & 77.03 \\
& & \checkmark & & 66.14 & 61.47 \\
& & \checkmark & \checkmark & 83.25 & 78.44 \\
& \checkmark & & & 99.79 & 97.68 \\
& \checkmark & & \checkmark & \textbf{99.82} & 97.51 \\
& \checkmark & \checkmark & & 99.79 & 99.29 \\
& \checkmark & \checkmark & \checkmark & 99.81 & 97.34 \\
\checkmark & & & & 98.27 & 96.69 \\
\checkmark & & & \checkmark & 97.73 & 97.21 \\
\checkmark & & \checkmark & & 98.28 & 96.73 \\
\checkmark & & \checkmark & \checkmark & 97.62 & 97.12 \\
\checkmark & \checkmark & & & 99.79 & 97.73 \\
\checkmark & \checkmark & & \checkmark & 99.81 & 97.78 \\
\checkmark & \checkmark & \checkmark & & 99.79 & 97.71 \\
\checkmark & \checkmark & \checkmark & \checkmark & \textbf{99.82} & \textbf{97.86} \\
\bottomrule
\end{tabular}
\end{table}

To quantify the contribution of each one of the features ($\mathcal{R}^{M\uparrow}$, $\mathcal{D}^{M\uparrow}$, $\mathcal{R}^{M\downarrow}$, $\mathcal{D}^{M\downarrow}$) on the NNIF method performance, we conducted an ablation study on CIFAR-10 dataset. Table~\ref{ablation_table} shows the AUC and accuracy results for Deepfool attack using features from the DNN's embedding space only. In the supp. material we present an extended ablation study with more attacks: FGSM, JSMA, and CW. 

Our analysis shows that the most influential feature is $\mathcal{D}^{M\uparrow}$, which is the $L_2$ distance from the most helpful training examples on the embedding space. In most cases, our NNIF detector performance using $\mathcal{D}^{M\uparrow}$ is nearly as good as the performance upon utilizing all four features. The least important feature is $\mathcal{R}^{M\downarrow}$, which barely helps the adversarial detection.
Intuitively it makes sense because we have noticed that the classes of the most harmful training examples always differ from the normal examples' class and mostly differ from the adversarial examples' class, and thus their rankings ($\mathcal{R}^{M\downarrow}$) are expected to be high for both cases (normal/adversarial). On the other hand, the distances from the most harmful training examples ($\mathcal{D}^{M\downarrow}$) are beneficial for the detection.
The most helpful ranks ($\mathcal{R}^{M\uparrow}$) is a beneficial feature when used by itself, alas incorporating it with $\mathcal{D}^{M\uparrow}$ did not improve the results. We therefore deduce that the information added by $\mathcal{R}^{M\uparrow}$ can already be inferred from $\mathcal{D}^{M\uparrow}$ in our detector.

We also show that the features $\mathcal{R}^{M\uparrow}$, $\mathcal{D}^{M\uparrow}$, $\mathcal{D}^{M\downarrow}$ affect every attack differently. We calculated the probability density functions for these three features on  CIFAR-10, applying the Deepfool and CW attacks (shown in the supp. material). From these histograms it can be easily observed that $\mathcal{R}^{M\uparrow}$ or $\mathcal{D}^{M\uparrow}$ are more useful for detecting Deepfool adversarial attacks than CW ones. On the other hand, the $\mathcal{D}^{M\downarrow}$ feature discriminates CW attacks better than Deepfool attacks.

A deployment of any learning based detector on systems is risky since an attacker could potentially have access to the LR classifier's parameters. Thus, it is helpful to deploy instead a detector which inspects only one feature and applies a simple thresholding. Our results show that this scheme is possible with NNIF using only the $\mathcal{D}^{M\uparrow}$ feature for all attacks.

\subsection{Generalization to other attacks}
To evaluate how well our detection method can be transferred to unseen attacks, we  trained LR classifiers on the features obtained using the FGSM attack, and then evaluated the classifies on the other (unseen) attacks. The AUC scores are shown in Table~\ref{generalization_table}. It can be observed that our NNIF method shows the best generalization everywhere except to JSMA. Table~\ref{generalization_table} results were collected using only the penultimate layer in the DNN (the embedding vector); A similar generalization table with additional attacks, using all the DNN layers, is provided in the supp. material. Notice that the generalization is weaker for all methods in this case. 

\begin{table}[h!]
\centering
\caption{Generalization of adversarial detection from FGSM attack to unseen attacks. The LR classifier is trained on the features extracted after applying FGSM attack, and then evaluated on JSMA, Deepfool, CW, PGD, and EAD.}
\label{generalization_table}
\resizebox{\columnwidth}{!}{
    \begin{tabular}{cc|cccccc}
    \toprule
    \multirow{2}{*}{Dataset} &  \multirow{2}{*}{Detector} & FGSM & \multirow{2}{*}{JSMA} & \multirow{2}{*}{Deepfool} & \multirow{2}{*}{CW} & \multirow{2}{*}{PGD} & \multirow{2}{*}{EAD} \\
    & & (seen) & & & & & \\
    \hline
    \multirow{4}{*}{CIFAR-10}  & D$k$NN      & 87.81     & 94.89          & 95.21          & 96.76          & 85.10 & 83.28 \\
                               & LID         & 90.12     & 94.67          & 95.43          & 97.66          & 90.29 & 82.52 \\
                               & Mahalanobis & 96.80     & \textbf{98.95} & 95.03          & 89.57          & 91.39 & 68.87 \\
                               & NNIF (ours) & 87.75     & 94.81          & \textbf{97.98} & \textbf{98.98} & \textbf{93.94} & \textbf{86.95} \\
    \hline
    \multirow{4}{*}{CIFAR-100} & D$k$NN      & 93.65     & 83.16          & 62.41          & 92.22          & 73.60 & 62.67 \\
                               & LID         & 80.68     & 74.33          & 52.25          & 67.84          & 72.25 & 52.10 \\
                               & Mahalanobis & 83.90     & \textbf{90.20} & 59.96          & 68.72          & 69.42 & 59.34 \\
                               & NNIF (ours) & 87.23     & 80.76          & \textbf{78.82} & \textbf{93.16} & \textbf{81.87} & \textbf{70.49} \\
    \hline
    \multirow{4}{*}{SVHN} .    & D$k$NN      & 85.24     & 93.43          & 89.84          & 92.20          & 75.99 & 79.81 \\
                               & LID         & 88.38     & 93.93          & 91.32          & 94.22          & 80.26 & 84.24 \\
                               & Mahalanobis & 98.14     & \textbf{99.00} & 91.46          & 87.51          & 86.26 & 80.62 \\
                               & NNIF (ours) & 91.06     & 97.91          & \textbf{95.79} & \textbf{98.16} & \textbf{89.80} & \textbf{91.99} \\
    \bottomrule
    \end{tabular}
    }
\end{table}

\subsection{Attack against NNIF}
\label{Adaptive attack against NNIF}
Here we consider a white-box threat model. In this setting, the adversary knows more than just the model parameters. We assume that the attacker is familiar with the adversarial defense scheme, but does not have access to the detector's parameters. Since the NNIF algorithm utilizes the entire training set, these data are also accessible to the attacker in our white-box setting. We employ a similar attack strategy as was proposed in \cite{Carlini2017BypassingTen} to evade the KD-based detector, and define a modified objective for the CW minimization:
\begin{equation}
\label{Opt}
\text{minimize } \norm{x - x_{adv}}_2^2 + c\cdot\left(\ell_{cw}(x_{adv}) + \ell_*(D(x_{adv}))\right),
\end{equation}
where $\ell_{cw}$ is the original adversarial loss term used in \cite{CarliniWagner2017Towards},
and $D(x_{adv})$ is the sum over all the distances (in the embedding space) between the adversarial image and the original image's most helpful training samples ($D_{adv}^+$ in Alg.~\ref{alg:NNIF}). More rigorously, we define:
\begin{equation}
\label{D_xadv}
\begin{split}
&\ell_*\left(D(x_{adv})\right) := \sum {D_{adv}^+} = \\
&\sum_{i=1}^M {\norm{\text{DNN}(x_{adv}) - \text{DNN}\left(X_{train}(H_{inds}^+[i])\right)}}_1,
\end{split}
\end{equation}
where DNN($\cdot$) is the network transformation from the input image to the embedding vector in the penultimate layer, and $H_{inds}^+[i]$ is the index of the $i^{th}$ most helpful training sample. Lastly, $c$ is a constant which balances between the fidelity to the original image and the adversarial strength.

The objective of the minimization in Eq.~\eqref{Opt} is to apply the CW attack while keeping $x_{adv}$ close to the most helpful training samples of the original image. In theory, we should have demanded this proximity for the nearest neighbors of $x_{adv}$, and not for $x_{adv}$ itself, but differentiating over the nearest neighbor algorithm is not feasible. It should also be noted that this attack was performed only on the penultimate activation layer, with the features that correspond to the most helpful examples: $\mathcal{R}^{M\uparrow}$ and $\mathcal{D}^{M\uparrow}$.

We applied this white-box attack on 4000 random samples of CIFAR-10 test set. We show the performances of D$k$NN, LID, Mahalanobis, and NNIF detectors on the original CW compared to our CW-Opt attack in Table~\ref{white_box}. Results for other datasets are in the supp. material. For every detector we used the same hyper-parameters which yielded the best defense results using only the last layer. Following \cite{Carlini2017BypassingTen}, we present the results in this experiment in term of accuracy, instead of AUC used in previous tests.

From Table~\ref{white_box} we observe that the proposed white-box attack decreases NNIF detection accuracy by only $1\%$. Therefore, we conclude that our NNIF defense algorithm is robust to a white-box setting. In addition, we note that the new attack impairs all the defense algorithms which rely on $L_2$ distance of nearest neighbors in the embedding space: D$k$NN, LID, and NNIF. Yet, for Mahalanobis we observe an adverse effect. This is somewhat expected since Mahalanobis estimates a global Gaussian for each class, and does not consider local features in the embedding space.

\begin{table}
\centering
\caption{Defense accuracy (\%) for a white-box attack targeting the NNIF detector on CIFAR-10.}
\label{white_box}
\begin{tabular}{c c c c c c}
\toprule
Attack & D$k$NN & LID & Mahalanobis & NNIF \\
\hline
CW     & 93.45 & 91.43 & 90.70 & 91.95 \\
CW-Opt & 90.99 & 89.74 & 92.29 & 90.81 \\
\bottomrule
\end{tabular}
\end{table}

\section{Discussion and conclusions}
\label{Discussion and conclusions}
In this paper, we addressed the task of detecting adversarial attacks. We showed that for normal (untempered) images, there exists a strong correlation between their nearest neighbors in the DNN's embedding space and their most helpful training examples, found using influence functions. Our empirical results show that the $L_2$ distance from a test image embedding vector to its most helpful training inputs ($\mathcal{D}^{M\uparrow}$) is a strong measure for the detection of adversarial examples. The aforementioned distance combined with the nearest neighbors ranking order of the training inputs were used to achieve a SOTA adversarial detection performance for six attacks (FGSM, JSMA, Deepfool, CW, PGD, EAD) on three datasets: CIFAR-10, CIFAR-100, and SVHN. Furthermore, we showed that our detector is robust in a white-box setting.

One possible avenue for future research is to inspect how the nearest neighbors are correlated with the most helpful/harmful training examples using different distance metrics or by employing a transform on the DNN embedding vectors. We emphasize that we mainly used the $L_2$ distance throughout our analysis, thus, we suspect that using another distance metric such as Mahalanobis \cite{Mahalanobis_adv_detection}
could improve our results further.

Another open issue for future research is the long computation time, which is required to calculate the influence functions for the entire training set. It is obvious that in order to deploy our NNIF algorithm, a significant improvement in computation time is needed, especially for real time applications or systems, which mandate fast detection pace. A possible solution to this problem may be a form of hash map from the nearest neighbors to the most influence training examples. Every training example can be encoded with a probability vector for its influence on a specific class; then, instead of employing a simple $k$-NN search in the embedding space, we can average over the probability of each class.

{\bf Acknowledgements.}
GS is partially supported by ARO, NGA, ONR, NSF, and gifts from Amazon, Google, and Microsoft.
RG and GC are supported by ERC-StG grant no. 757497 (SPADE) and gifts from NVIDIA, Amazon, and Google.

{\small
\bibliographystyle{ieee_fullname}
\bibliography{my_bib}

\begin{thebibliography}{10}\itemsep=-1pt

\bibitem{DBLP:journals/corr/BahdanauCB14}
Dzmitry Bahdanau, Kyunghyun Cho, and Yoshua Bengio.
\newblock Neural machine translation by jointly learning to align and
  translate.
\newblock {\em arXiv:1409.0473}, 2014.

\bibitem{OnEvaluatingAdvRobustness2019}
Nicholas Carlini, Anish Athalye, Nicolas Papernot, Wieland Brendel, Jonas
  Rauber, Dimitris Tsipras, Ian Goodfellow, Aleksander Madry, and Alexey
  Kurakin.
\newblock On evaluating adversarial robustness.
\newblock {\em arXiv:1902.06705}, 2019.

\bibitem{Carlini2017BypassingTen}
Nicholas Carlini and David~A Wagner.
\newblock Adversarial examples are not easily detected: Bypassing ten detection
  methods.
\newblock In {\em AISec@CCS}, 2017.

\bibitem{CarliniWagner2017Towards}
Nicholas Carlini and David~A Wagner.
\newblock Towards evaluating the robustness of neural networks.
\newblock {\em 2017 IEEE Symposium on Security and Privacy (SP)}, pages 39--57,
  2017.

\bibitem{EAD}
Pin-Yu Chen, Yash Sharma, Huan Zhang, Jinfeng Yi, and Cho-Jui Hsieh.
\newblock Ead: Elastic-net attacks to deep neural networks via adversarial
  examples.
\newblock In {\em AAAI}, 2018.

\bibitem{Parseval}
Moustapha Cisse, Piotr Bojanowski, Edouard Grave, Yann Dauphin, and Nicolas
  Usunier.
\newblock Parseval networks: Improving robustness to adversarial examples.
\newblock In {\em ICML}, 2017.

\bibitem{DNN_or_kNN}
Gilad Cohen, Guillermo Sapiro, and Raja Giryes.
\newblock {DNN} or k-{NN}: That is the generalize vs. memorize question.
\newblock {\em arXiv:1805.06822}, 2018.

\bibitem{Doring2017_knn_convergence}
Maik D{\"{o}}ring, László Gy{\"{o}}rfi, and Harro Walk.
\newblock Rate of convergence of k-nearest-neighbor classification rule.
\newblock {\em J. Mach. Learn. Res.}, 18(1):8485--8500, 1 2017.

\bibitem{Dubey2019DefenseAA}
Abhimanyu Dubey, Laurens van~der Maaten, Zeki Yalniz, Yixuan Li, and
  Dhruv~Kumar Mahajan.
\newblock Defense against adversarial images using web-scale nearest-neighbor
  search.
\newblock {\em CVPR}, pages 8759--8768, 2019.

\bibitem{ClassificationRegions}
Alhussein Fawzi, Seyed-Mohsen Moosavi-Dezfooli, Pascal Frossard, and Stefano
  Soatto.
\newblock {Classification regions of deep neural networks}.
\newblock {\em arXiv:1705.09552}, 2017.

\bibitem{Feinman2017DetectingAS}
Reuben Feinman, Ryan~R Curtin, Saurabh Shintre, and Andrew~B Gardner.
\newblock Detecting adversarial samples from artifacts.
\newblock {\em arXiv:1703.00410}, 2017.

\bibitem{FGSM}
Ian Goodfellow, Jonathon Shlens, and Christian Szegedy.
\newblock Explaining and harnessing adversarial examples.
\newblock In {\em ICLR}, 2015.

\bibitem{RESNET}
Kaiming He, Xiangyu Zhang, Shaoqing Ren, and Jian Sun.
\newblock Deep residual learning for image recognition.
\newblock {\em CVPR}, pages 770--778, 2016.

\bibitem{Cross_Lipschitz_regularization}
Matthias Hein and Maksym Andriushchenko.
\newblock Formal guarantees on the robustness of a classifier against
  adversarial manipulation.
\newblock In {\em NIPS}, 2017.

\bibitem{Hinton2012DeepRecognition}
Geoffrey Hinton, Li Deng, Dong Yu, George Dahl, Abdel-Rahman Mohamed, Navdeep
  Jaitly, Andrew Senior, Vincent Vanhoucke, Patrick Nguyen, Tara Sainath, and
  Brian Kingbury.
\newblock Deep neural networks for acoustic modeling in speech recognition.
\newblock {\em IEEE Signal Processing Magazine}, 29(6):82--97, 2012.

\bibitem{Jakubovitz2018ImprovingDR}
Daniel Jakubovitz and Raja Giryes.
\newblock Improving dnn robustness to adversarial attacks using jacobian
  regularization.
\newblock In {\em ECCV}, 2018.

\bibitem{to_trust_or_not_a_classifier}
Heinrich Jiang, Been Kim, Melody Guan, and Maya Gupta.
\newblock To trust or not to trust a classifier.
\newblock In {\em NIPS}, pages 5546--5557, 2018.

\bibitem{Kim2014ConvolutionalClassification}
Yoon Kim.
\newblock Convolutional neural networks for sentence classification.
\newblock In {\em EMNLP}, pages 1746--1751, 2014.

\bibitem{Koh2017UnderstandingBP}
Pang~Wei Koh and Percy~S. Liang.
\newblock Understanding black-box predictions via influence functions.
\newblock In {\em ICML}, 2017.

\bibitem{CIFAR}
Alex Krizhevsky.
\newblock Learning multiple layers of features from tiny images.
\newblock Technical report, 2009.

\bibitem{Krizhevsky2012ImageNetNetworks}
Alex Krizhevsky, Ilya Sutskever, and Geoffrey~E Hinton.
\newblock Imagenet classification with deep convolutional neural networks.
\newblock {\em Advances In Neural Information Processing Systems}, pages 1--9,
  2012.

\bibitem{Kurakin2017AdversarialML}
Alexey Kurakin, Ian~J. Goodfellow, and Samy Bengio.
\newblock Adversarial machine learning at scale.
\newblock {\em arXiv:1611.01236}, 2017.

\bibitem{Mahalanobis_adv_detection}
Kimin Lee, Kibok Lee, Honglak Lee, and Jinwoo Shin.
\newblock A simple unified framework for detecting out-of-distribution samples
  and adversarial attacks.
\newblock {\em NeurIPS}, 2018.

\bibitem{Li2017AdversarialED}
Xin Li and Fuxin Li.
\newblock Adversarial examples detection in deep networks with convolutional
  filter statistics.
\newblock {\em ICCV}, pages 5775--5783, 2017.

\bibitem{LID}
Xingjun Ma, Bo Li, Yisen Wang, Sarah~M Erfani, Sudanthi N~R Wijewickrema,
  Michael~E Houle, Grant Schoenebeck, Dawn Song, and James Bailey.
\newblock Characterizing adversarial subspaces using local intrinsic
  dimensionality.
\newblock {\em arXiv:1801.02613}, 2018.

\bibitem{Towards_DeepLearning_Resistance}
Aleksander Madry, Aleksandar Makelov, Ludwig Schmidt, Dimitris Tsipras, and
  Adrian Vladu.
\newblock Towards deep learning models resistant to adversarial attacks.
\newblock In {\em ICLR}, 2018.

\bibitem{Meng2017MagNetAT}
Dongyu Meng and Hao Chen.
\newblock Magnet: A two-pronged defense against adversarial examples.
\newblock In {\em ACM}, 2017.

\bibitem{Metzen17detecting}
Jan~Hendrik Metzen, Tim Genewein, Volker Fischer, and Bastian Bischoff.
\newblock On detecting adversarial perturbations.
\newblock In {\em ICLR}, 2017.

\bibitem{Virtual_Adversarial_Training}
Takeru Miyato, Shin-ichi Maeda, Masanori Koyama, Ken Nakae, and Shin Ishii.
\newblock Distributional smoothing with virtual adversarial training.
\newblock In {\em ICLR}, 2015.

\bibitem{Dezfooli17Universal}
Seyed-Mohsen Moosavi-Dezfooli, Alhussein Fawzi, Omar Fawzi, and Pascal
  Frossard.
\newblock Universal adversarial perturbations.
\newblock In {\em CVPR}, 2017.

\bibitem{Curvature}
Seyed-Mohsen Moosavi-Dezfooli, Alhussein Fawzi, Omar Fawzi, Pascal Frossard,
  and Stefano Soatto.
\newblock Analysis of universal adversarial perturbations.
\newblock {\em ArXiv:1705.09554}, 2017.

\bibitem{DeepFool}
Seyed-Mohsen Moosavi-Dezfooli, Alhussein Fawzi, and Pascal Frossard.
\newblock Deepfool: A simple and accurate method to fool deep neural networks.
\newblock {\em CVPR}, pages 2574--2582, 2016.

\bibitem{SVHN}
Yuval Netzer, Tao Wang, Adam Coates, Alessandro Bissacco, Bo Wu, and Andrew~Y
  Ng.
\newblock Reading digits in natural images with unsupervised feature learning.
\newblock 2011.

\bibitem{cleverhans}
Nicolas Papernot, Fartash Faghri, Nicholas Carlini, Ian Goodfellow, Reuben
  Feinman, Alexey Kurakin, Cihang Xie, Yash Sharma, Tom Brown, Aurko Roy,
  Alexander Matyasko, Vahid Behzadan, Karen Hambardzumyan, Zhishuai Zhang,
  Yi-Lin Juang, Zhi Li, Ryan Sheatsley, Abhibhav Garg, Jonathan Uesato, Willi
  Gierke, Yinpeng Dong, David Berthelot, Paul Hendricks, Jonas Rauber, and
  Rujun Long.
\newblock Technical report on the cleverhans v2.1.0 adversarial examples
  library.
\newblock {\em arXiv:1610.00768}, 2018.

\bibitem{Distillation}
N. {Papernot}, P. {McDaniel}, X. {Wu}, S. {Jha}, and A. {Swami}.
\newblock Distillation as a defense to adversarial perturbations against deep
  neural networks.
\newblock In {\em IEEE Symposium on Security and Privacy (SP)}, 2016.

\bibitem{Deep_kNN_Papernot}
Nicolas Papernot and Patrick~D McDaniel.
\newblock Deep k-nearest neighbors: Towards confident, interpretable and robust
  deep learning.
\newblock {\em arXiv:1803.04765}, 2018.

\bibitem{JSMA}
Nicolas Papernot, Patrick~D McDaniel, Somesh Jha, Matt Fredrikson, Z~Berkay
  Celik, and Ananthram Swami.
\newblock The limitations of deep learning in adversarial settings.
\newblock {\em IEEE European Symposium on Security and Privacy (EuroS{\&}P)},
  pages 372--387, 2016.

\bibitem{InputGradients}
Andrew~Slavin Ross and Finale Doshi-Velez.
\newblock Improving the adversarial robustness and interpretability of deep
  neural networks by regularizing their input gradients.
\newblock In {\em AAAI}, 2017.

\bibitem{Rouhani2018TowardsSD}
Bita~Darvish Rouhani, Mohammad Samragh, Tara Javidi, and Farinaz Koushanfar.
\newblock Towards safe deep learning: Unsupervised defense against generic
  adversarial attacks.
\newblock 2018.

\bibitem{BANG}
Andras Rozsa, Manuel Gunther, and Terrance E.~Boult.
\newblock Towards robust deep neural networks with bang.
\newblock In {\em WACV}, 2018.

\bibitem{Schroff2015FaceNet:Clustering}
Florian Schroff, Dmitry Kalenichenko, and James Philbin.
\newblock {FaceNet: A} unified embedding for face recognition and clustering.
\newblock In {\em CVPR}, 2015.

\bibitem{Shaham2018RobustOptimization}
Uri Shaham, Yutaro Yamada, and Sahand Negahban.
\newblock {Understanding adversarial training: Increasing local stability of
  supervised models through robust optimization}.
\newblock {\em Neurocomputing}, 307:195--204, 2018.

\bibitem{Vulnerability_Input_Dimension}
Carl-Johann Simon-Gabriel, Yann Ollivier, Léon Bottou, Bernhard
  Sch{\"{o}}lkopf, and David Lopez-Paz.
\newblock First-order adversarial vulnerability of neural networks and input
  dimension.
\newblock In {\em ICML}, 2019.

\bibitem{CertifyingRobustness}
Aman Sinha, Hongseok Namkoong, and John Duchi.
\newblock Certifiable distributional robustness with principled adversarial
  training.
\newblock In {\em ICLR}, 2018.

\bibitem{Sitawarin2019DefendingAA}
Chawin Sitawarin and D{\'a}vid W{\'a}gner.
\newblock Defending against adversarial examples with k-nearest neighbor.
\newblock {\em ArXiv}, abs/1906.09525, 2019.

\bibitem{Intriguing}
Christian Szegedy, Wojciech Zaremba, Ilya Sutskever, Joan Bruna, Dumitru Erhan,
  Ian Goodfellow, and Rob Fergus.
\newblock Intriguing properties of neural networks.
\newblock In {\em ICLR}, 2014.

\bibitem{Transferable}
Florian Tramer, Nicolas Papernot, Ian Goodfellow, Dan Boneh1, and Patrick
  McDaniel.
\newblock The space of transferable adversarial examples.
\newblock {\em arXiv:1704.03453}, 2017.

\bibitem{tramer2018ensemble}
Florian Tramèr, Alexey Kurakin, Nicolas Papernot, Ian Goodfellow, Dan Boneh,
  and Patrick McDaniel.
\newblock Ensemble adversarial training: Attacks and defenses.
\newblock In {\em ICLR}, 2018.

\bibitem{Voulodimos2018DeepReview}
Athanasios Voulodimos, Nikolaos Doulamis, Anastasios Doulamis, and Eftychios
  Protopapadakis.
\newblock Deep learning for computer vision: A brief review.
\newblock {\em Computational Intelligence and Neuroscience}, 2018.

\bibitem{Xu2018FeatureSD}
Weilin Xu, David Evans, and Yanjun Qi.
\newblock Feature squeezing: Detecting adversarial examples in deep neural
  networks.
\newblock {\em arXiv:1704.01155}, 2018.

\bibitem{DBLP:journals/corr/ZhangPBZLBC17}
Ying Zhang, Mohammad Pezeshki, Philemon Brakel, Saizheng Zhang, César Laurent,
  Yoshua Bengio, and Aaron~C Courville.
\newblock Towards end-to-end speech recognition with deep convolutional neural
  networks.
\newblock In {\em Interspeech}, 2016.

\end{thebibliography}
}

\onecolumn
\appendix
\section{Method}
\label{supp_Method}
The main paper proposes a new reactive detection method for adversarial images: the Nearest Neighbors Influence Functions (NNIF). Our detector utilizes a influence functions algorithm as shown in \cite{Koh2017UnderstandingBP} to measure the contribution of each training sample to a test samples prediction. Their algorithm is summarized in Algorithm~\ref{alg:InfluenceFunction}. For measuring the influence a train sample $z$ has on the loss of a specific test sample $z_{test}$, \cite{Koh2017UnderstandingBP} approximate this term in Eq.~\eqref{I_up_loss}:
\begin{equation*}
I_{up,loss}(z, z_{test}) = -\nabla_{\theta}L(z_{test}, \theta)^TH_{\theta}^{-1}\nabla_{\theta}L(z, \theta),
\end{equation*}
where $H$ is the Hessian of the machine learning model, $L$ is its loss, and $\theta$ are the model parameters.
Eq.~\eqref{I_up_loss} is repeated throughout the training set, calculating $I_{up,loss}$ for every training sample. For our NNIF algorithm only the top $M$ helpful training examples ($H_{inds}^+$) and the top $M$ harmful training examples ($H_{inds}^-$) are chosen for further processing.

\begin{algorithm}[H]
\caption{Influence Functions}\label{alg:InfluenceFunction}
\begin{algorithmic}[1]
\Require Test sample $(x_i, y_i$) and a training set ($X_{train}$, $Y_{train}$)
\Require $M$: Number of top influence samples to collect
\Ensure $H_{inds}^+$, $H_{inds}^-$ \Comment{Most helpful/harmful training examples indices}
\State $N_{train} = |X_{train}|$
\State Initialize $H_{inds}^+$=[], $H_{inds}^-$=[] 
\State Initialize $I_{up,loss}$ = zeros[$N_{train}$]
\For{($x_j, y_j$) in ($X_{train}$, $Y_{train}$)}
\State $I_{up,loss}[j] = -\nabla_{\theta}L(x_i, \theta)^TH_{\theta}^{-1}\nabla_{\theta}L(x_j, \theta)$ \Comment{Apply influence function (Eq.~\eqref{I_up_loss})}
\EndFor
\State sort($I_{up,loss}[j]$) \Comment{Sorting for the most influential training samples}
\For{$m$ in $[0,M-1]$}
    \State $j_{m}^+$ = Training example index of $I_{up,loss}[N_{train}-m]$ \Comment{choosing most helpful examples}
    \State $H_{inds}^+$.append($j_{m}^+$)
    \State $j_{m}^-$ = Training example index of $I_{up,loss}[m]$ \Comment{choosing most harmful examples}
    \State $H_{inds}^-$.append($j_{m}^-$)
\EndFor
\State \textbf{return} $H_{inds}^+$, $H_{inds}^-$ \Comment{Most helpful/harmful training examples indices}
\end{algorithmic}
\end{algorithm}

\clearpage
\section{Experimental setup}
The DNNs clean accuracies, when not under attack, are shown in Table~\ref{clean_acc}. In Table~\ref{attack_rates} we present the attack success rate of the Fast Gradient Sign Method (FGSM) (\cite{FGSM}), Jacobian-based Saliency Map Attack (JSMA) (\cite{JSMA}), Deepfool (\cite{DeepFool}), Carlini \& Wagner (CW) (\cite{CarliniWagner2017Towards}), our CW-Opt attack, Projected Gradient Descent (PGD) (\cite{Towards_DeepLearning_Resistance}), and Elastic-net Attack on Dnns (EAD) (\cite{EAD}). Note that the success rates of all attacks are higher for CIFAR-100. This makes sense since CIFAR-100 dataset has 100 classes instead of 10, and it is thus more vulnerable to misclassifications.

\begin{table}[h!]
\centering
\caption{DNN clean accuracies (\%), for normal images not under attack.}
\label{clean_acc}
\begin{tabular}{c|ccc}
\toprule
Dataset      & train acc & val acc & test acc \\
\hline
CIFAR-10     & 99.75     & 93.70    & 92.08 \\
CIFAR-100    & 96.80     & 70.80    & 67.99 \\
SVHN         & 99.46     & 96.20    & 94.59 \\
\bottomrule
\end{tabular}
\end{table}

\begin{table*}[h!]
\centering
\caption{Adversarial attack success rates (\%) of FGSM, JSMA, Deepfool, CW, CW-Opt, PGD, and EAD. CW-Opt attack is CW regulated with a loss term optimized against our NNIF defense in a white-box setting.}
\label{attack_rates}
\resizebox{\linewidth}{!}{
  \begin{tabular}{c|cc|cc|cc|cc|cc|cc|cc}
  \toprule
  \multirow{2}{*}{Dataset} & \multicolumn{2}{c|}{FGSM} & \multicolumn{2}{c|}{JSMA} & \multicolumn{2}{c|}{Deepfool} & \multicolumn{2}{c|}{CW} & \multicolumn{2}{c|}{CW-Opt} & \multicolumn{2}{c|}{PGD} & \multicolumn{2}{c}{EAD} \\
  & val & test & val & test & val & test & val & test & val & test & val & test & val & test \\
  \hline
  CIFAR-10  & 80.47 &	79.27 &	71.18 &	70.21 &	94.34  & 96.19 & 93.70 & 94.46 & 86.87 & 86.31 & 79.62 & 80.51 & 46.64 & 48.14 \\
  CIFAR-100 & 95.19 & 95.26 &	86.02 &	86.19 &	100.00 & 99.91 & 99.44 & 98.90 & 99.15 & 99.10 & 99.58 & 99.25 & 86.86 & 89.41 \\
  SVHN      & 84.72 &	85.51 &	69.02 &	65.51 &	92.62  & 92.45 & 93.24 & 95.69 & 49.69 & 45.96 & 39.09 & 47.73 & 75.99 & 77.44 \\
  \bottomrule
  \end{tabular}
}
\end{table*}

The paper explains how we tuned the hyper-parameters for the four inspected algorithms: D$k$NN, LID, Mahalanobis, and our NNIF method. For the D$k$NN and LID algorithms we tuned the number of neighbors ($k$), for the Mahalanobis algorithm we tuned the noise magnitude ($\epsilon$), and for our NNIF method we set the number of top influence samples to collect ($M$). All parameters were chosen using nested cross entropy validation within the validation set, based on the AUC values of the detection ROC curve. The results are shown in Table~\ref{parameters}.

\begin{table}[h!]
\centering
\caption{Hyper-parameter setting for the four inspected detectors. $k$ denotes the number of nearest neighers used in D$k$NN and LID algorithms, $\epsilon$ is the noise magnitude in the Mahalanobis detector, and $M$ is the number of most helpful/harmful training images used in our NNIF method.}
\label{parameters}
\resizebox{\columnwidth}{!}{
    \begin{tabular}{cc|cccccc}
    \hline
    Dataset & Param & FGSM & JSMA & Deepfool & CW & PGD & EAD \\
    \hline 
    \multirow{4}{*}{CIFAR-10} & D$k$NN ($k$)    & 4900 & 5000 & 4900 & 4900 & 4800 & 4900 \\
              & LID ($k$)                       & 18   & 18   & 18   & 18   & 24 & 16 \\
              & Mahalanobis ($\epsilon$)        & 0.0002 & 0.0002 & 0.00005 & 0.00001 & 0.00005 & 0.00001 \\
              & NNIF ($M$)                      & 50   & 200 & 100 & 200 & 450 & 500 \\ 
    \hline
    \multirow{4}{*}{CIFAR-100} & D$k$NN ($k$)   & 490  & 450  & 20 & 430 & 500 & 10 \\
              & LID ($k$)                       & 10   & 10   & 10 & 10 & 10 & 10 \\
              & Mahalanobis ($\epsilon$)        & 0.005 & 0.005 & 0.0005 & 0.00001 & 0.01 & 0.0002 \\
              & NNIF ($M$)                      & 30   & 30   & 40 & 40 & 50 & 30 \\ 
    \hline
    \multirow{4}{*}{SVHN} & D$k$NN ($k$)        & 3200 & 3000 & 1400 & 3200 & 3200 & 3200 \\
              & LID ($k$)                       & 18   & 22   & 22 & 22 & 22 & 24 \\
              & Mahalanobis ($\epsilon$)        & 0.001 & 0.0005  & 0.00005 & 0.00001 & 0.00008 & 0.00001 \\
              & NNIF ($M$)                      & 300 & 50    & 300  & 50 & 100 & 100 \\ 
    \hline
    \end{tabular}
    }
\end{table}

\clearpage
\section{Detection of adversarial attacks}
\label{supp: Detection of adversarial attacks}
\begin{figure*}[h!]
\centering
\includegraphics[width=\linewidth]{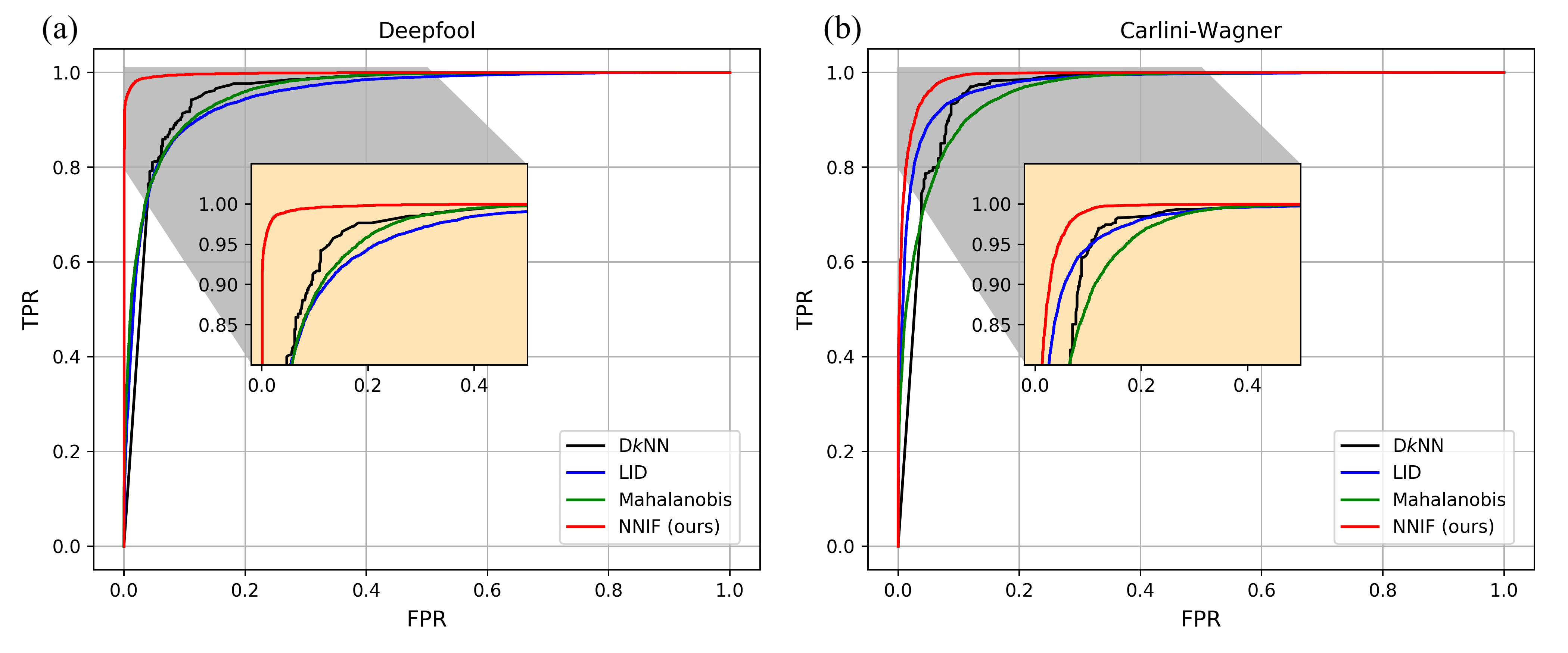}
\caption{ROC curves for classifying adversarial examples. (a) Defending Deepfool attack. (b) Defending Carlini-Wagner (CW) $L_2$ attack. All plots correspond to the CIFAR-10 dataset. We achieve state-of-the-art results, surpassing previous defense methods by a large margin.}
\label{images/auc_cifar10}
\end{figure*}

Figure~\ref{images/auc_cifar10} presents two ROC curves for classification of Deepfool and CW adversarial attacks on the CIFAR-10 dataset. One can observe that our NNIF method (solid red line) achieves better classification power over the previous state-of-the-art methods.

Table~\ref{detection_scores_last_layer} presents the AUC scores for the adversarial detection of FGSM, JSMA, Deepfool, CW, PGD, and EAD attacks on CIFAR-10, CIFAR-100, and SVHN datasets. These results were obtained by using DNN's features from only the embedding space. A similar table with detectors which were trained on the entire DNN's features is in the main paper.

\begin{table}[h!]
\centering
\caption{Comparison of AUC scores (\%) for various adversarial detection methods, for FGSM, JSMA, Deepfool, CW, PGD, and EAD attacks. Results obtained using only the DNN's penultimate layer.}
\label{detection_scores_last_layer}
\resizebox{\columnwidth}{!}{
    \begin{tabular}{cc|cccccc}
    \toprule
    Dataset & Detector & FGSM & JSMA & Deepfool & CW & PGD & EAD\\
    \hline
    \multirow{4}{*}{CIFAR-10}  & D$k$NN      & 87.81          & 95.37          & 95.82          & 96.88          & 86.83 & 85.20 \\
                               & LID         & 90.12          & 94.67          & 95.43          & 97.66          & 90.49 & 82.87 \\
                               & Mahalanobis & \textbf{96.80} & \textbf{98.95} & 96.49          & 96.96          & 92.91 & 85.30 \\ 
                               & NNIF (ours) & 87.75          & 97.67          & \textbf{99.82} & \textbf{99.05} & \textbf{94.01} & \textbf{88.06} \\
    \hline
    \multirow{4}{*}{CIFAR-100} & D$k$NN      & \textbf{93.65} & 83.46          & 76.71          & 93.77          & 73.78 & \textbf{78.42} \\
                               & LID         & 80.68          & 74.33          & 52.25          & 67.84          & 72.25 & 52.10 \\
                               & Mahalanobis & 83.90          & \textbf{90.20} & 62.05          & 71.60          & 72.46 & 61.65 \\ 
                               & NNIF (ours) & 87.23          & 86.63          & \textbf{84.20} & \textbf{94.58} & \textbf{83.09} & 72.42 \\
    \hline
    \multirow{4}{*}{SVHN}      & D$k$NN      & 85.24          & 94.61          & 91.13          & 95.15          & 79.07 & 84.77 \\
                               & LID         & 88.38          & 94.31          & 92.00          & 95.64          & 80.92 & 86.74 \\
                               & Mahalanobis & \textbf{98.14} & \textbf{99.15} & 96.07          & 98.26          & 90.41 & 92.95 \\
                               & NNIF (ours) & 91.06          & 98.29          & \textbf{97.11} & \textbf{98.68} & \textbf{92.46} & \textbf{93.72} \\
    \bottomrule
    \end{tabular}
    }
\end{table}

\clearpage
\section{Ablation study}
To inspect how the four learned features influence our adversarial detection we conducted an ablation study on CIFAR-10 dataset, for four attacks: FGSM, JSMA, Deepfool, and CW. The results are shown in Table~\ref{Supp_ablation_table}. From these results, one may conclude that the most beneficial feature is $\mathbb{D}^{M\uparrow}$, which is the $L_2$ distance from the most helpful training examples on the deep neural network (DNN) embedding space.

\begin{figure}[t!]
\centering
\includegraphics[width=\linewidth]{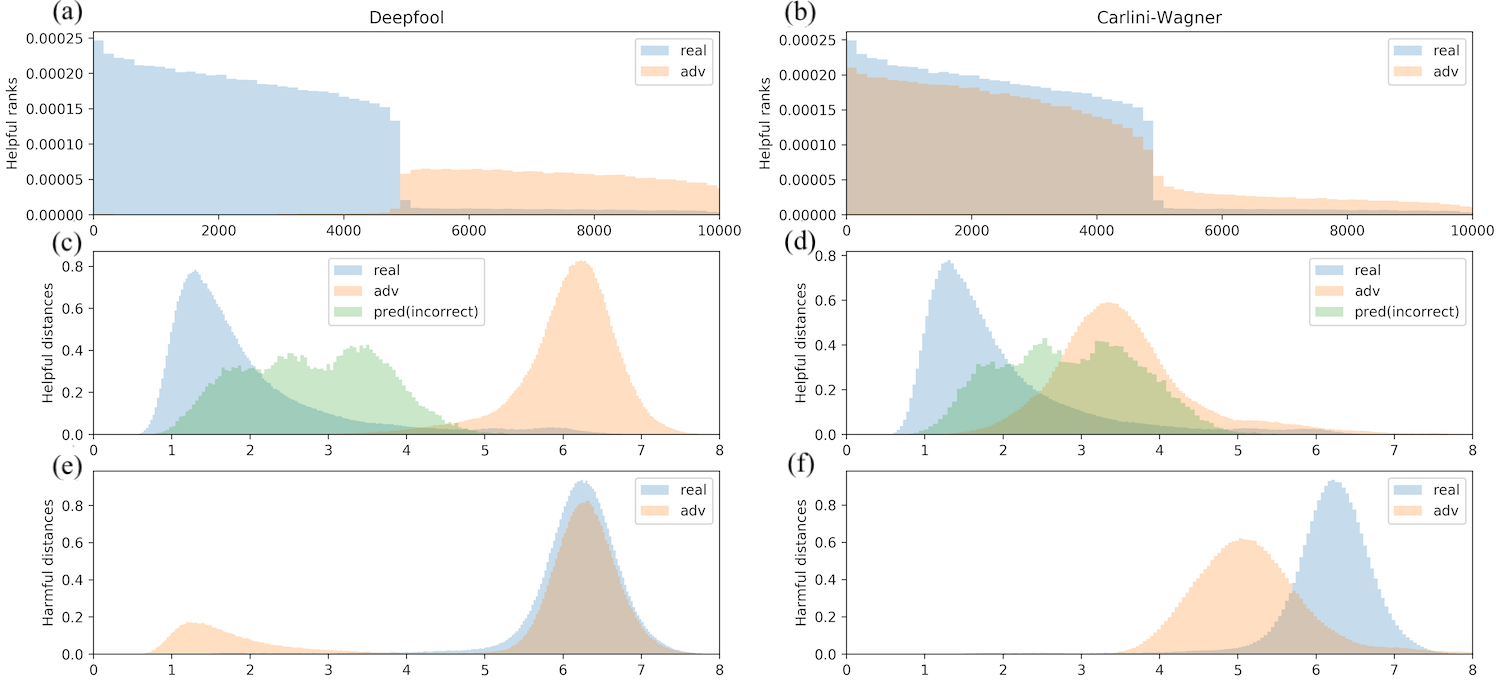}
\caption{Probability density functions of the most helpful ranks ($\mathbb{R}^{M\uparrow}$, top row), most helpful distances ($\mathbb{D}^{M\uparrow}$, middle row), and the most harmful distances ($\mathbb{D}^{M\downarrow}$, bottom row), on CIFAR-10 for the Deepfool and CW attacks. The features for the normal (untempered) images that were correctly classified by the network are shown in blue. The features for the adversarial images are shown in orange. The features for the normal images that were misclassified by the network are shown in green (in the middle row).}
\label{images/histograms_full}
\end{figure}

Figure~\ref{images/histograms_full} shows the probability density functions for $\mathbb{R}^{M\uparrow}$, $\mathbb{D}^{M\uparrow}$, and $\mathbb{D}^{M\downarrow}$ features on CIFAR-10 for the Deepfool and CW adversarial attacks. From these histograms, it can be easily observed that $\mathbb{R}^{M\uparrow}$ or $\mathbb{D}^{M\uparrow}$ are more useful for detecting Deepfool adversarial attacks than CW attacks. On the other hand, the $\mathbb{D}^{M\downarrow}$ feature discriminates CW attacks better than Deepfool attacks. This is also supported by the results on Table~\ref{Supp_ablation_table}: For $\mathbb{R}^{M\uparrow}$ or $\mathbb{D}^{M\uparrow}$ alone NNIF detects Deepfool better than CW ($98.27\% > 81.91\%$ and $99.79\% > 97.27\%$), however, for $\mathbb{D}^{M\downarrow}$ NNIF is able to detect CW attacks better than Deepfool attacks ($89.97\% > 82.11\%$).

\begin{table}[h!]
\centering
\caption{Ablation test for adversarial attack detection: Calculating AUC score and accuracy for selected features. Attacking CIFAR-10 dataset using FGSM, JSMA, Deepfool, and CW.}
\label{Supp_ablation_table}
\resizebox{\columnwidth}{!}{
    \begin{tabular}{cccc|cccc}
    \toprule
    $\mathbb{R}^{M\uparrow}$ & $\mathbb{D}^{M\uparrow}$ & $\mathbb{R}^{M\downarrow}$ & $\mathbb{D}^{M\downarrow}$ & FGSM & JSMA & Deepfool & CW \\
    \hline
    & & & \checkmark & 78.99 & 83.23 & 82.11 & 89.97 \\
    & & \checkmark & & 51.4  & 51.93 & 66.14 & 53.14 \\
    & & \checkmark & \checkmark & 82.08 & 85.11 & 83.25 & 90.27 \\
    & \checkmark & & & 84.19 & 97.41 & 99.79 & 97.27 \\
    & \checkmark & & \checkmark & 86.74 & 97.54 & 99.82 & 98.81 \\
    & \checkmark & \checkmark & & 84.20 & 97.41 & 99.79 & 97.27 \\
    & \checkmark & \checkmark & \checkmark & 87.74 & 97.66 & 99.81 & 99.0 \\
    \checkmark & & & & 64.85 & 85.27 & 98.27 & 81.91 \\
    \checkmark & & & \checkmark & 80.19 & 85.4 & 97.73 & 95.14 \\
    \checkmark & & \checkmark & & 64.31 & 85.34 & 98.28 & 81.95 \\
    \checkmark & & \checkmark & \checkmark & 83.14 & 85.97 & 97.62 & 95.34 \\
    \checkmark & \checkmark & & & 84.18 & 97.43 & 99.79 & 97.21 \\
    \checkmark & \checkmark & & \checkmark & 86.66 & 97.51 & 99.81 & 98.85 \\
    \checkmark & \checkmark & \checkmark & & 84.22 & 97.44 & 99.79 & 97.21 \\
    \checkmark & \checkmark & \checkmark & \checkmark & \textbf{87.75} & \textbf{97.67} & \textbf{99.82} & \textbf{99.05} \\
    \bottomrule
    \end{tabular}
}
\end{table}

\clearpage
\section{Generalization to other attacks}
The main paper measures the NNIF method transferability from one attack (FGSM) to other, unseen attacks (JSMA, Deepfool, CW, PGD, and EAD), where all the features are extracted from the penultimate activation layer. Here we provide a similar table where all the DNN's activation layers are employed for this comparison (Table~\ref{Supp_generalization_table}), except of D$k$NN which only utilizes features from the DNN's embedding space. The generalization results in Table~\ref{Supp_generalization_table} does not have a definite winner method. The D$k$NN, Mahalanobis, and our NNIF methods demonstrate the best transferability for various setups. The LID detector shows the worst generalization overall.

\begin{table}[h!]
\centering
\caption{Generalization of adversarial detection from FGSM attack to unseen attacks. The LR classifier is trained on all activation layers' features extracted after applying FGSM attack, and then evaluated on JSMA, Deepfool, CW, PGD, and EAD.}
\label{Supp_generalization_table}
\resizebox{\columnwidth}{!}{
    \begin{tabular}{cc|cccccc}
    \toprule
    \multirow{2}{*}{Dataset} &  \multirow{2}{*}{Detector} & FGSM & \multirow{2}{*}{JSMA} & \multirow{2}{*}{Deepfool} & \multirow{2}{*}{CW} & \multirow{2}{*}{PGD} & \multirow{2}{*}{EAD} \\
    & & (seen) & & & & & \\
    \hline
    \multirow{4}{*}{CIFAR-10}  & D$k$NN      & 87.81     & 94.89          & \textbf{95.21} & \textbf{96.76} & 85.10 & \textbf{83.28} \\
                               & LID         & 98.18     & 91.70          & 84.51          & 91.67          & \textbf{85.62} & 70.85 \\
                               & Mahalanobis & 99.80     & \textbf{96.11} & 86.25          & 85.17          & 84.24 & 68.30 \\
                               & NNIF (ours) & 99.96     & 92.76          & 79.84          & 84.44          & 81.66 & 70.02 \\
    \hline
    \multirow{4}{*}{CIFAR-100} & D$k$NN      & 93.65     & 83.16          & 62.41          & \textbf{92.22} & 73.60 & 62.67 \\
                               & LID         & 92.33     & 72.65          & 51.19          & 59.09          & 64.49 & 51.00 \\
                               & Mahalanobis & 99.87     & 82.26          & 52.15          & 53.72          & 52.94 & 52.58 \\
                               & NNIF (ours) & 99.96     & \textbf{89.52} & \textbf{64.33} & 86.43          & \textbf{85.79} & \textbf{63.64} \\
    \hline
    \multirow{4}{*}{SVHN}      & D$k$NN      & 85.24     & 93.43          & 89.84          & \textbf{92.20} & 75.99 & 79.81 \\
                               & LID         & 99.92     & 94.91          & 82.55          & 82.26          & 69.90 & 73.40 \\
                               & Mahalanobis & 100.00    & \textbf{99.18} & \textbf{92.24} & 86.87          & \textbf{82.57} & \textbf{81.06} \\
                               & NNIF (ours) & 100.00    & 92.45          & 80.14          & 83.20          & 75.74 & 75.52 \\
    \bottomrule
    \end{tabular}
    }
\end{table}

\clearpage
\section{Attack against NNIF}
We applied a white-box attack against our NNIF defense model on CIFAR-10/100 and SVHN datasets, CW-Opt (Section 4.5 in the main paper). This attack optimization requires a hyper-parameter in the new regularization term, $M$. This is the number of the most helpful training examples of the normal image. We apply this term only on the top $1\%$ helpful training samples which belong to the predicted class (we find this to be most effective for the attack to succeed). Therefore, we set $M=50$ for CIFAR-10 and SVHN and $M=5$ for CIFAR-100. Table~\ref{white_box} shows the D$k$NN, LID, Mahalanobis, and our NNIF detection accuracies on two scenarios: 1) With the vanilla CW attack and 2) With our white-box attack (CW-Opt).

\begin{table}[h!]
\centering
\caption{Attack failure rate without defense (\%) and defense accuracy (\%) for a white-box attack targeting the NNIF detector. The attack failure rate in the third column corresponds to the probability of the adversary to fail flipping a correct label without any defense method.}
\label{supp_white_box}
\begin{tabular}{c c c| c c c c}
\toprule
Dataset & Attack & Attack fail rate & \multicolumn{4}{c}{Defense accuracy (\%)} \\
& & (w.o. defense) (\%) & D$k$NN & LID & Mahalanobis & NNIF \\
\hline
\multirow{2}{*}{CIFAR-10}  & CW     & 5.54  & 93.45 & 91.43 & 90.70 & 91.95 \\
                           & CW-Opt & 13.69 & 90.99 & 89.74 & 92.29 & 90.81 \\
\hline
\multirow{2}{*}{CIFAR-100} & CW     & 1.10  & 87.42 & 61.37 & 64.16 & 85.42 \\
                           & CW-Opt & 0.90  & 94.16 & 66.05 & 51.98 & 91.15 \\
\hline               
\multirow{2}{*}{SVHN}      & CW     & 4.31  & 91.03 & 87.91 & 93.24 & 94.65 \\
                           & CW-Opt & 54.04 & 65.59 & 70.21 & 77.23 & 75.21 \\
\bottomrule
\end{tabular}
\end{table}

For CIFAR-10 we observe only a $1\%$ decrease in our NNIF adversarial detection accuracy. Similar decrease is present also for all the algorithms which utilize $L_2$ distance of nearest neighbors in the embedding space: D$k$NN and LID.

For SVHN we observe that CW-Opt attack impairs our NNIF defense by $20\%$. We speculate this is because CW-Opt was able to flip only $46\%$ of labels in the SVHN test set, instead of $96\%$ where attacking with the vanilla CW. Therefore, in the white-box setting we consider only the hardest test samples for our detection task. We also notice that D$k$NN and LID defense accuracies are decreased by more than $20\%$ as well.

The results for CIFAR-100 are unconformable to the other datasets, showing an increase of the NNIF detection accuracy in the white-box setting. This finding also presents with D$k$NN and LID, which is correlative to the trend shown on CIFAR-10. This happens since the attack focuses only on the most helpful distance feature and our defense takes into account also other parameters. Therefore, to verify our white-box attack indeed brings an adversarial image closer to its natural image's helpful training images (in the embedding space), we repeated the experiment by only collecting the distance features, $\mathcal{D}^{M\uparrow}$, in our defense and ignoring the ranks, $\mathcal{R}^{M\uparrow}$. This method demonstrates a decrease of the detection accuracy from $74\%$ to $65\%$. This shows that indeed the white box attack also affects CIFAR-100 when it relies only on distance features.   The detection accuracies using only $\mathcal{D}^{M\uparrow}$ are summarized in Table~\ref{white_box_only_D}.  Note that our defense technique is always robust to the white-box attacks when it only uses the distance features. The fact that we show robustness also when considering the ranks features makes it even stronger since it is hard to optimize the white-box attacks to ranks (as they are non-differentiable).

Overall, we conclude that our NNIF defense method is robust in a white-box setting.

\begin{table}[h!]
\centering
\caption{Defense accuracy (\%) for a white-box attack targeting the NNIF detector, using only the distance features $\mathcal{D}^{M\uparrow}$.}
\label{white_box_only_D}
\begin{tabular}{c c c}
\toprule
Dataset & Attack & NNIF defense acc. \\
\hline
\multirow{2}{*}{CIFAR-10}  & CW     & 91.96 \\
                           & CW-Opt & 90.91 \\
\hline
\multirow{2}{*}{CIFAR-100} & CW     & 74.09 \\
                           & CW-Opt & 65.48 \\
\hline               
\multirow{2}{*}{SVHN}      & CW     & 94.65 \\
                           & CW-Opt & 75.27 \\
\bottomrule
\end{tabular}
\end{table}

\clearpage
\section{Influence function smoothness}
Since we use ReLU activations in our Resnet-34 DNN, the cross entropy loss function is not continuously differentiable, therefore we might have an issue calculating the influence function in Eq.~\eqref{I_up_loss}. Although this is a technical concern, in practice we can assume this is not an issue since the set of discontinuities has measure zero, and the problematic activation points will never be encountered in the back propagation.


\end{document}